\documentclass[letterpaper, 10 pt, conference]{ieeeconf}  

\IEEEoverridecommandlockouts                              

\usepackage{graphics} 
\usepackage{epsfig} 
\usepackage{amsmath} 
\usepackage{amssymb}  
\usepackage{xcolor,colortbl}
\usepackage{subfigure}
\usepackage{hyperref}
\usepackage{bbm}

\usepackage[nolist]{acronym}
\usepackage[textsize=tiny]{todonotes}
\usepackage{placeins}
\usepackage{bigstrut}
\usepackage{soul}
\usepackage{subfigure}
\usepackage{gensymb}
\usepackage{caption}
\usepackage{booktabs}

\DeclareMathOperator*{\argmax}{argmax}

\DeclareMathOperator*{\E}{\mathbb{E}}

\newcommand\maxpage[2][Error]{  
\ifnum\value{page}>#2
    \gderror{\Large \bf On page {\thepage} we are past page #2 (too long).   #1 }
\else\fi
}

\title{\LARGE \bf
Learning to Drive Off Road on Smooth Terrain in Unstructured Environments Using an On-Board Camera and Sparse Aerial Images
}

\author{Travis Manderson, Stefan Wapnick, David Meger, Gregory Dudek%
\thanks{Mobile Robotics Laboratory, 
	School of Computer Science,
	McGill University, Montreal, Canada\newline
        {\tt\small \{travism,swapnick,dmeger,dudek\}@cim.mcgill.ca}}%
}

\newcommand{\norm}[1]{\left\lVert#1\right\rVert}
\newcommand{\etal}{\textit{et al.}}

\begin{document}
\setlength{\abovecaptionskip}{0.5pt}
\setlength{\belowcaptionskip}{-8pt}
\setlength{\abovedisplayskip}{3pt}
\setlength{\belowdisplayskip}{3pt}

\setlength{\textfloatsep}{10pt}

\bstctlcite{MyBSTcontrol}


\maketitle
\thispagestyle{empty}
\pagestyle{empty}

\begin{abstract}

We present a method for learning to drive on smooth terrain while simultaneously avoiding collisions in challenging off-road and unstructured outdoor environments using only visual inputs. Our approach applies a hybrid model-based and model-free reinforcement learning method that is entirely self-supervised in labeling  terrain roughness and collisions using on-board sensors. Notably, we provide both first-person and overhead aerial image inputs to our model. We find that the fusion of these complementary inputs improves planning foresight and makes the model robust to visual obstructions. Our results show the ability to generalize to environments with plentiful vegetation, various types of rock, and sandy trails.   
During evaluation, our policy attained 90\% smooth terrain traversal and reduced the proportion of rough terrain driven over by 6.1 times compared to a model using only first-person imagery.
Video and project details can be found at
\url{www.cim.mcgill.ca/mrl/offroad\_driving/}.
\end{abstract}


\section{INTRODUCTION}
\label{sec_introduction}

\begin{figure}[t!]
  \centering
  \includegraphics[width=1 \linewidth]{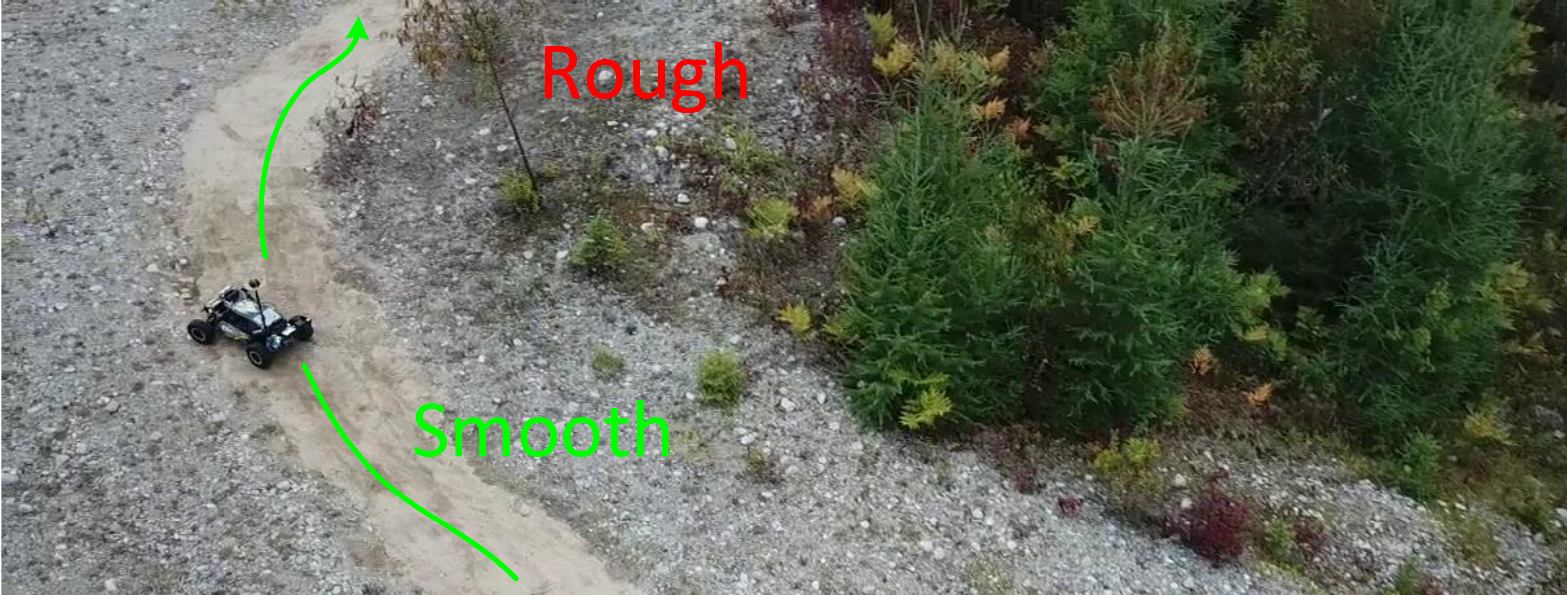}
  \caption{Example of smooth and rough terrain in an unstructured, outdoor environment.}
\label{fig:terrain-example}
\end{figure}

In this paper, we present a system for learning a navigation policy that preferentially chooses smooth terrain (see Fig.~\ref{fig:terrain-example}) for off-road driving in natural, unstructured environments using an on-board camera and sparse aerial images. The emphasis of the paper, however, is not 
road classification {\em per se}, but rather to propose an approach for online adaptive self-supervised learning for off-road driving in rough terrain and to explore the synthesis of aerial and first-person (ground) sensing in this context.

Scaled robot vehicles, such as remote-controlled vehicles, are capable of covering large distances at high speeds when on paved road, but compared to their larger counterparts, they are less capable of traversing rough terrain when navigating off road. However, their small size also makes them a good choice for operating in remote natural environments due to their limited disturbance and impact on the environment.

Modern \acp{MAV} have become ubiquitous due their ease of deployment, continually increasing endurance and range, and high quality cameras at a relatively low cost. Although lightweight \acp{MAV} lack the payload to carry heavy sensors, they are able to provide a top-down view of the surface and an unrestricted \ac{FOV} by either taking pictures at higher altitude or in a mosaic sequence. In this work, we use a single image taken at 80~m above the surface, combined with a \ac{RTK} \ac{GPS} to enable our off-road vehicle to better reason about its surroundings. This technique enhances the ability of a vehicle to do long range planning beyond its own ground-based \ac{FOV}.

Eventual large scale and long duration deployment of robots in remote, natural, unstructured, and complex environments will require these robots to be robust and to reason about unforeseen situations that they will inevitably encounter.
There has been extensive work on navigating outdoors \cite{Thrun2006}, \cite{Urmson2008}, \cite{Suger2015}. These methods generally rely on complex geometric and sensor modeling combined with terrain classification trained from labeled data, or they require human demonstrations.

Reinforcement learning methods - especially model-free methods that learn from vision - are rarely used in real-world scenarios due to high dimensionality and sample inefficiency~\cite{Dulac-Arnold2019}. Model-based approaches are more sample efficient but may perform poorly due to cascading model errors and the fact that the learned dynamics is only a proxy to determining a policy to maximize the agent's reward.

Inspired by the recent work by Kahn~\etal~\cite{Kahn2018}, we combine aspects of model-free and model-based methods into a single computation graph to leverage the strengths of both methods while offsetting their individual weaknesses. 
Our model learns to navigate collision-free trajectories while maximizing smooth terrain in a purely self-supervised fashion with limited training data. 
During training, terrain roughness is estimated using an on-board \ac{IMU} while obstacles are measured using a short-range Lidar. To achieve greater visual context, we fuse input images from an on-board first-person camera and local aerial view. We validate our system both in simulation and on a real off-road vehicle in a remote environment. We demonstrate the vehicle's ability to navigate smooth, collision-free trajectories and show that the addition of aerial imagery improves predictive performance, especially in the presence of sharp turning trajectories or visual obstructions.

\section{RELATED WORK}
\label{sec_related_work}

\subsection{Imitation Learning for Visual Navigation}

Learning vision-based controllers for autonomous driving has been studied for several decades. Pomerleau~\etal~\cite{pomerleau1989alvinn,jochem1993maniac} used behavioral cloning to train a feed-forward neural network to predict steering angles for on-road driving. Training images and Lidar data were collected in simulation, while the system was demonstrated to work on real roads in limited conditions. More recently, Bojarski~\etal~\cite{bojarski2016end} trained a \ac{DNN} to predict steering angles from a single forward-facing camera using human demonstration. Additional training images were collected from left-shifted and right-shifted cameras to depict recovery cases when the vehicle drifts off course. The labels of these images were adjusted to turn right and left, respectively.

Bearing some resemblance to our own work, several researchers have explored autonomous navigation in forests and unstructured environments by following natural or human-made paths. Rasmussen~\etal~\cite{Rasmussen2014} used an appearance and structural-based model from stereo cameras and Lidar to predict the most likely path. 
Some authors have treated path navigation as an image classification problem. Giusti~\etal~\cite{Giusti2016} and Smolyanskiy~\etal~\cite{Smolyanskiy2017} collected data using three offset cameras (left/center/right) while walking through forest trails. The images were labeled with desired yaw actions where the left-offset and right-offset cameras were used to depict situations when the learner drifts from the expert path (the forward-facing image labeled as zero yaw while the left and right images were labeled as yaw right and left, respectively). The \ac{DNN} was then validated using a \ac{MAV} to autonomously navigate forest trails.

In our previous work~\cite{Manderson2018iros}, we used behavioral cloning and an approach similar to DAGGER~\cite{ross2011reduction} to train a \ac{DNN} to predict yaw and pitch angles for an underwater robot. The behavioral objective was to navigate close to coral while simultaneously avoiding collisions and moving towards regions of interest. We iteratively labeled, trained, and evaluated our image-based controller during several deployments until the robot achieved acceptable performance.




\subsection{Semantic Segmentation for Terrain Traversability}

A common approach to visual navigation in the presence of diverse terrain is a two step process of visual semantic segmentation followed by geometric path planning. 
A number of authors have addressed urban navigation using such an approach. Barnes \etal~\cite{Barnes2017} and Tang \etal~\cite{Tang2017} trained a network to segment the agent's view into drivable, non-drivable, and unknown regions using labeled trajectories generated from expert demonstrations. More recent work by Wellhausen~\etal~\cite{Wellhausen2019} applied semantic segmentation in a self-supervised fashion by using a quadruped robot's recorded force-torque signals to label traversed terrain. Labeled terrain was projected into the robot's camera frame for generating a training set for segmentation. For path planning, the world was discretized into grid cells where cost was derived from the re-projected predicted segmented images and Dijkstra's algorithm \cite{dijkstra1959note} was used to find the optimal path.
For environments involving many terrain classes, Rothrock~\etal~\cite{Rothrock2016} applied the DeepLab~\cite{ChenPK0Y16} segmentation network with first-person to top-down image projection for the task of evaluating traversability for Mars rovers.

\subsection{Air-Assisted Ground Vehicle Navigation}
Several researchers have presented methods using aerial imagery to assist ground navigation. Sofman~\etal~\cite{Sofman2006} trained a classifier on aerial images of urban environments discretized into grid cells. The classification label of each cell was translated to a cost and used for D* planning~\cite{dstar} of a ground vehicle. 
Similarly, Delmerico~\etal~\cite{Delmerico2017}~\cite{Delmerico2017_2} proposed a lightweight, patch-based terrain classification network requiring only minimal training time from aerial images. The classifier output forms a cost map for D* planning for the purpose of guiding a ground vehicle in disaster recovery situations. Wang~\etal~\cite{Wang2018} used a team of air and ground vehicles for collaborative exploration to map an area more quickly. Garzon~\etal~\cite{Garzon2013} combined sensor data from air and ground vehicles to accurately perform obstacle pose estimation and used a potential fields-based approach for collision-free navigation.

\subsection{Sensor Fusion}

Sensor fusion encompasses various techniques for combining different input modalities to improve model performance. One common approach is to train an ensemble of models on different sensory inputs and use voting or measures of uncertainty to combine predictions. Lee~\etal~\cite{Lee2018} trained an ensemble of uncertainty-aware policy networks composed of a left camera, right camera, and GPS input modalities. They applied their ensemble to the task of autonomous driving where the network output possessing minimum uncertainty was chosen.
Other approaches directly input multiple sensory streams into the same network under the assumption that the model itself will learn to automatically weigh each input optimally~\cite{Chen2017}~\cite{Ku2018}~\cite{Engelcke2017}~\cite{Xu2018}. 

\begin{figure*}[t]
\vspace{0.07in}
  \centering
  \includegraphics[width=0.9\linewidth]{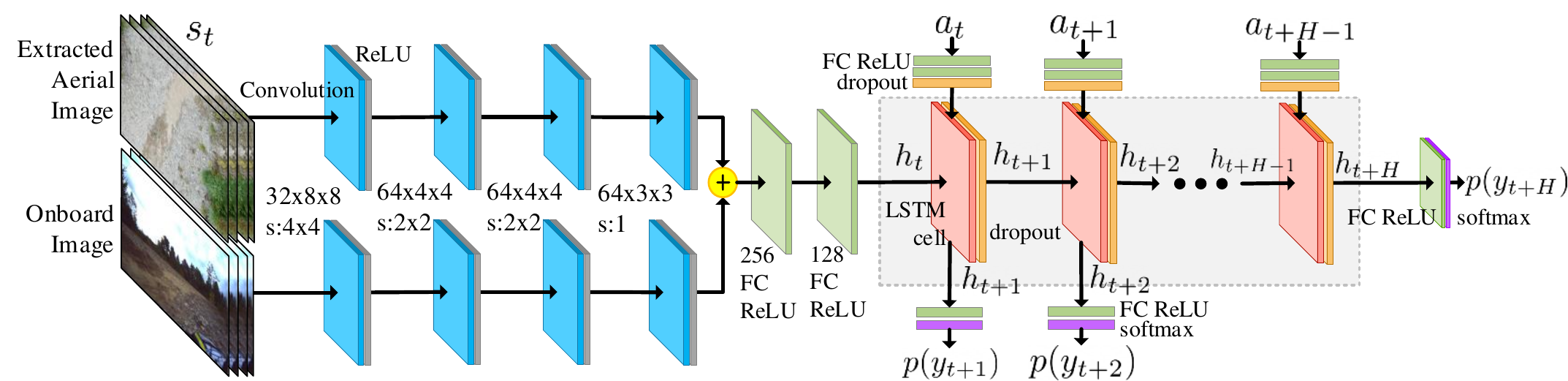}
  \caption{Neural network model for predicting terrain types over a horizon $H$. The state $s_t$ is composed of the current and several previous ground and aerial aligned images to increase context. The action inputs $a_t$ through $a_{t+H-1}$ are steering angles. The image inputs travel through the convolutional block to form the initial hidden state of the recurrent network. The softmax outputs $p(y_{t+i})$ represent the probability of each terrain class. Dropout is optionally used to reduce overfitting.}
\label{fig:network}
\end{figure*}

\section{PRELIMINARIES}
\label{sec_background}

\subsection{Reinforcement Learning}
In reinforcement learning, the environment is represented as a Markov Decision Process, $\mathcal{M}=(\mathcal{S}, \mathcal{A}, \mathcal{R}, \mathcal{T})$, representing the set of states, actions, rewards, and transition dynamics, respectively. The agent seeks to learn a control policy $a_t\sim\pi(s_t)$ that maximizes the return $G_t$: 

\begin{equation}
G_t = {\sum_{k=0}^T{\gamma}^{k}R_{t+k+1}}
\end{equation}
where $0 \leq \gamma \leq 1$ is the discount factor and $R_{k}$ is the reward.
The value function of a state $V_{\pi}(s_t)$ is defined as the expected return by following policy $\pi$ from state $s_t$. The value function can be defined iteratively by the Bellman equation: 
\begin{gather}
\begin{aligned}
\begin{split}
V_{\pi}(s_t) & = \E_{\pi}\left[ G_t \right]\\
& = \sum_{a_{t}}\pi(a_t|s_t)\sum_{s_{t+1}}p(s_{t+1}|s_t,a_t)  \left[r_{t}+{\gamma}V_{\pi}(s_{t+1})\right]
\end{split}
\end{aligned}\raisetag{3\baselineskip}
\end{gather}
where $p(s_{t+1}|s_t,a_t)$ represents the transition dynamics of the system.
In model-free reinforcement learning, an agent learns a control policy to directly maximize the return without explicitly learning the dynamics of the system. Conversely, in model-based reinforcement learning, the agent first seeks to learn the system dynamics and then plans a policy by simulating rollouts. Typically, the learned dynamics model takes the form of a state predictive function $s_{t+1} = f(s_{t},a_{t})$.

\subsection{Value Prediction Networks and Generalized Computation Graphs}
Recently, Oh~\etal~\cite{Oh2017} proposed Value Prediction Networks (VPN), a hybrid model-based and model-free reinforcement learning architecture. Value Prediction Networks learn an encoded state $h_{t}$ optimized to predict immediate rewards $r_t$ and value functions on the abstract state $V(h_{t})$. Planning is done by simulating rollouts on abstract states $h_{t}$ over a horizon of timesteps from which the action that maximizes the average expected return (calculated jointly from rewards $r_t$ and value functions $V(h_{t})$) is taken. The intuition behind this approach is that planning need not require full observation predictions (such as images) and can instead be done by only predicting the upcoming rewards and value functions on a learned model of encoded abstract states. Oh~\etal~\cite{Oh2017} assert that Value Predictive Networks can be viewed as jointly model-based and model-free. It is model-based as it implicitly learns a dynamics model for abstract states optimized for predicting future rewards and value functions. It is also model-free as it maps these encoded abstract states to rewards and value functions using direct experience with the environment prior to the planning phase.

Oh~\etal~\cite{Oh2017} found that this hybrid model possessed several advantages over conventional model-free and model-based learning methods. Specifically, they found it to be more sample efficient and to generalize better than Deep Q-Networks~\cite{MnihKSGAWR13} while simultaneously outperforming a conventional observation predictive network in cases where planning on the raw state space is difficult due to the space's complexity or the environment is highly stochastic.


Khan~\etal~\cite{Kahn2018} developed a similar architecture, known as Generalized Computation Graphs (GCG), for the task of collision avoidance over a short predictive horizon. Their model was deployed on a small, remote-controlled car in an indoor corridor environment. They found that predicting discrete reward values as in a classification problem (e.g. collision or no collision), instead of real-valued outputs, improved performance. Furthermore, they found that removing the prediction on the bootstrapped value function improved learning efficiency and stability without hindering performance because this real-valued, recursive quantity can be more difficult to predict and may not yield noticeable performance gains for a short-horizon control task.

\section{MODEL OVERVIEW} 
\label{sec:model_overview}

\subsection{Representation}
\label{representation}

We propose a deep learning model for predicting the terrain roughness and collision probabilities over a fixed horizon of $H$ timesteps from which planning can be done to derive an action policy. We use an architecture similar to Khan~\etal~\cite{Kahn2018}, which has previously been shown to produce good performance in short-term control situations. The model operates on an input image state $s_t$ and action sequence of steering commands $\langle a_t, a_{t+1},... a_{t+H-1} \rangle$. The image state is composed of the current image and recent visual history of the last M timesteps ($s_t = \langle I_{t-M+1}, ..., I_t \rangle$) to provide additional context. For this input sequence, the model predicts the probability each terrain classes over the planning horizon $\langle p(y_{t+1}), p(y_{t+2}),...,p(y_{t+H}) \rangle$ where $y_t \in C$ and $C$ is the set of all terrain classes. We choose increasing label values to be rougher terrain (0 being completely smooth and $|C|-1$ being an obstacle). Throttle is assumed to be constant and therefore is not present in the action space.

This architecture can be seen as modeling the joint probability of terrain classes $\langle y_{t+1}, y_{t+2}...y_{t+H} \rangle$ over a horizon $H$ while assuming conditional independence between labels:
\begin{equation}\label{eq:prob_representation}
\begin{split}
p(y_{t+H}, ... y_{t+1}|a_{t+H-1}...a_{t}, s_{t}) =\\ 
\prod_{i=1}^{H}{p(y_{t+i}|a_{t+i-1}...a_{t}, s_{t})}
\end{split}
\end{equation}
In the context of Value Prediction Networks, the predicted terrain labels act as reward values. To enforce higher rewards for more desirable or smoother terrain classes, we remap the label to a reward value: 
\begin{equation}\label{eq:reward_mapping}
r_t = |C|-1-y_{t}
\end{equation}

\subsection{Network Architecture and Objective Function}
\label{network-architecture}
The network architecture modeling the mapping from an image state and action sequence to predicted terrain class probabilities is shown in Fig. \ref{fig:network}. The current image state $s_t$ is put through a convolutional neural network to form the initial hidden state of a Long Short-Term Memory (LSTM)~\cite{Hochreiter1997} recurrent neural network. Air and ground image inputs follow different convolutional branches as their learned features differ based on their unique perspectives. The feature map outputs of these dual branches are concatenated. Steering actions $a_t$ and predicted terrain class probabilities $p(\hat{y}_{t+1})$ form the input-output pair of the recurrent network.

The cross-entropy loss objective for training is obtained by applying maximum log likelihood to Eq. \ref{eq:prob_representation} with L2 regularization:
\begin{equation}\label{eq:loss_function}
L_{t} = -\sum_{i=1}^{H} \sum_{c_{j}\in{C}} \textbf{1}_{y_{t+i} = c_{j}} \log{{p(\hat{y}_{t+i}|a_{t+i-1}...a_{t}, s_{t})}} + \lambda\norm{w}_{2}^{2}
\end{equation}
where $y_{t+i}$ and $\hat{y}_{t+i}$ are the true and predicted labels.

To instantiate our network, we choose $M=4$ as the visual history length to give $s_t = \langle I_{t-3}, I_{t-2}, I_{t-1}, I_{t} \rangle$ and use color image inputs of size of 72x128x3. The visual image history inputs are concatenated as additional channels, making the overall image input size 72x128x12 for both air and ground. Four convolutional layers consisting of (32, 64, 64, 64) kernel units of sizes (8x8, 4x4, 4x4, 3x3) and of strides (4, 2, 2, 1) were chosen, respectively. Additional parameters unique to each experimentation environment (simulation and real-world) are detailed in section \ref{sec:experimental_results}.




\subsection{Planning}
 \label{planning}
During the planning phase, we seek to solve for a trajectory of actions that maximizes the expected reward over a planning horizon of $H$ (where rewards follow Eq. \ref{eq:reward_mapping}): 
\begin{equation}
A_{t}^{H} = \argmax_{a_t, ...a_{t+H-1}} -\sum_{i=1}^{H}\sum_{c_{j}\in C}c_{j}*{p(\hat{y}_{t+i}=c_{j}|a_{t+i-1}...a_{t}, s_{t})}
\end{equation}

To solve this optimization, a randomized K-shooting method is applied~\cite{tedrake2009underactuated}. K action rollouts are randomly generated and each k'th ($1\leq k\leq K$) rollout $A_{(t,k)}^{H}=(a_t, ...a_{t+H-1})_k$ is evaluated using our learned model to predict the expected cumulative trajectory reward. Our policy applies the first action from the the trajectory with the highest perceived return. Following the Model Predictive Control paradigm, planning is repeated at every time step.

\section{SYSTEM OVERVIEW}
\label{sec_system_overview}

\begin{figure}[t!]
\vspace{0.07in}
  \centering
  \includegraphics[width=0.9\linewidth]{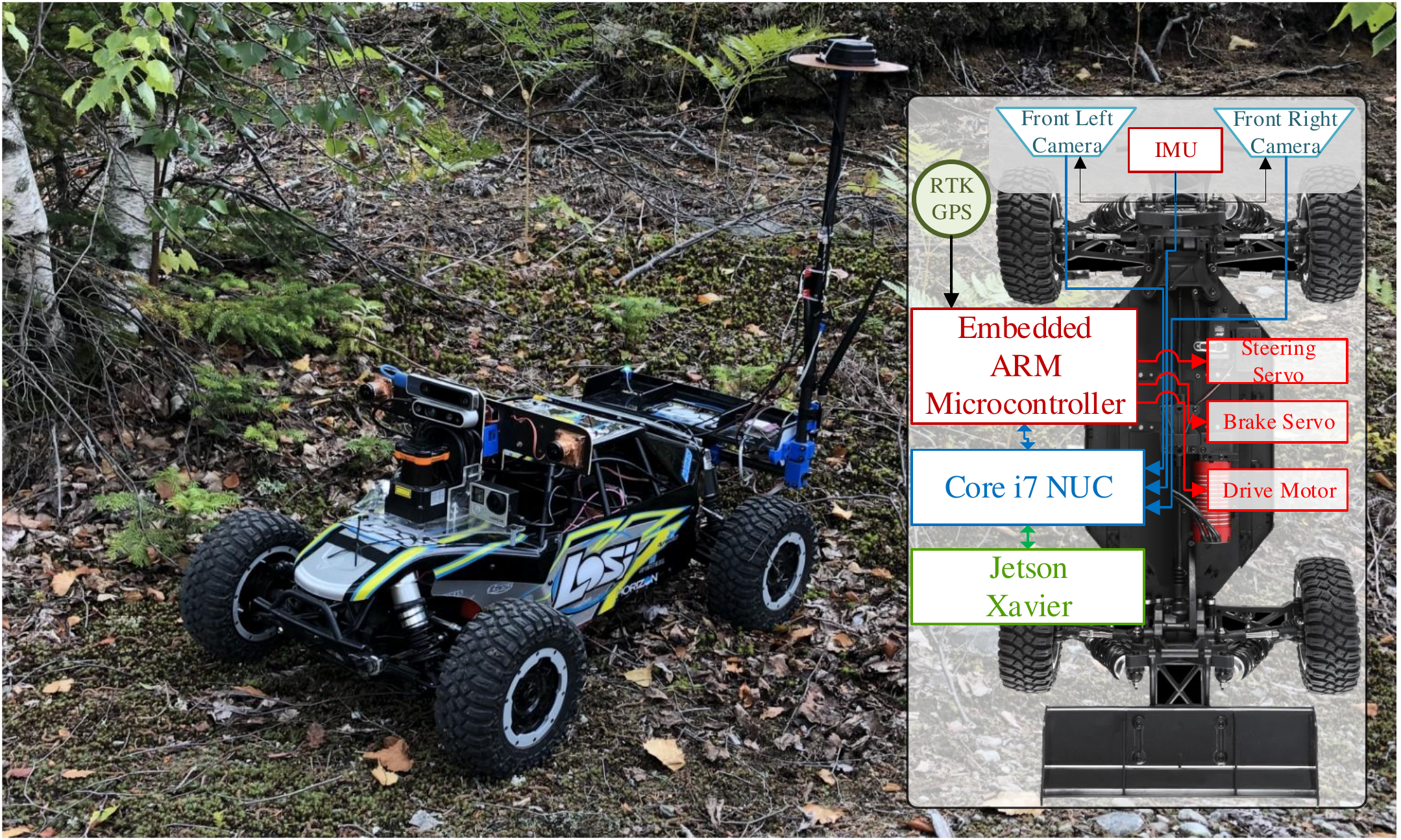}
  \caption{Off-road vehicle used for validation experiments. 
  Only one forward-facing camera is used for these experiments.}
\label{fig:vehicle}
\end{figure}

Our experimental off-road vehicle is based on a modified 1:5 scale, electric off-road, remote-controlled buggy as shown in Fig.~\ref{fig:vehicle}. It contains two servo motors for controlling the steering and mechanical brake, while a 260~kV sensored brushless motor drives the four wheels. An embedded ARM microcontroller maintains a pose estimate using an extended Kalman filter that fuses sensor information from redundant triple-axis accelerometers, magnetometers, gyros, and a u-blox ZED-F9P \ac{RTK} \ac{GPS}. The microcontroller also controls the two servo motors and brushless motor controller by forwarding commands either from a user hand-held remote or from a connected Intel i7 NUC computer when autonomously driven. The microcontroller receives \ac{RTK} correction signals through a long-range wireless link and relays all sensor information to the NUC. Mounted on the front of the vehicle is a Hokuo UTM-30LX Lidar and two forward-facing IDS Imaging UI-3251LE global-shutter cameras with 4mm lenses which are tightly coupled to a VectorNav VN-100 \ac{IMU} (although only one camera was used in our experiments).

The Intel i7 NUC mini PC runs Ubuntu and the \ac{ROS} software package. The NUC records camera images, Lidar obstacles detections, and \ac{IMU} readings. An NVIDIA Jetson Xavier is also connected to the NUC via a shared gigabit network. The Xavier likewise runs the \ac{ROS} software package and hosts our terrain predictive neural network model (Fig.~\ref{fig:network}) implemented in Tensorflow~\cite{abadi2016tensorflow}.

Before deploying the ground vehicle, sparse aerial images are taken using a \ac{MAV} at a high altitude above the deployment region. In our case, a single image taken at 80~m provided a resolution of 0.01~meters-per-pixel. Using four landmarks measured with the ground vehicle, we shift, rotate, and scale the image to align with the landmarks. Visual inspection showed this image to be accurate within 0.1~m.
At each timestep during runtime, we use the GPS location and compass heading to extract a 12~m~x~9~m aerial image patch that is oriented with the ground vehicle and centered 1.5~m directly in front of the vehicle.

Many authors have considered the use of an \ac{IMU} to measure road roughness~\cite{Kertesz2007,wen2008road,barsi2006potential}. Similarly, we measure the \ac{RMS} linear acceleration reported by the robot's \ac{IMU} over short time windows to approximate an instantaneous roughness score of the traversed terrain.


\section{EXPERIMENTS}
\label{sec:experimental_results}

\vspace*{-0.1in}
\subsection{Simulation}

\begin{figure}[t]
\vspace{0.07in}
  \centering
  \includegraphics[width=1.0\linewidth]{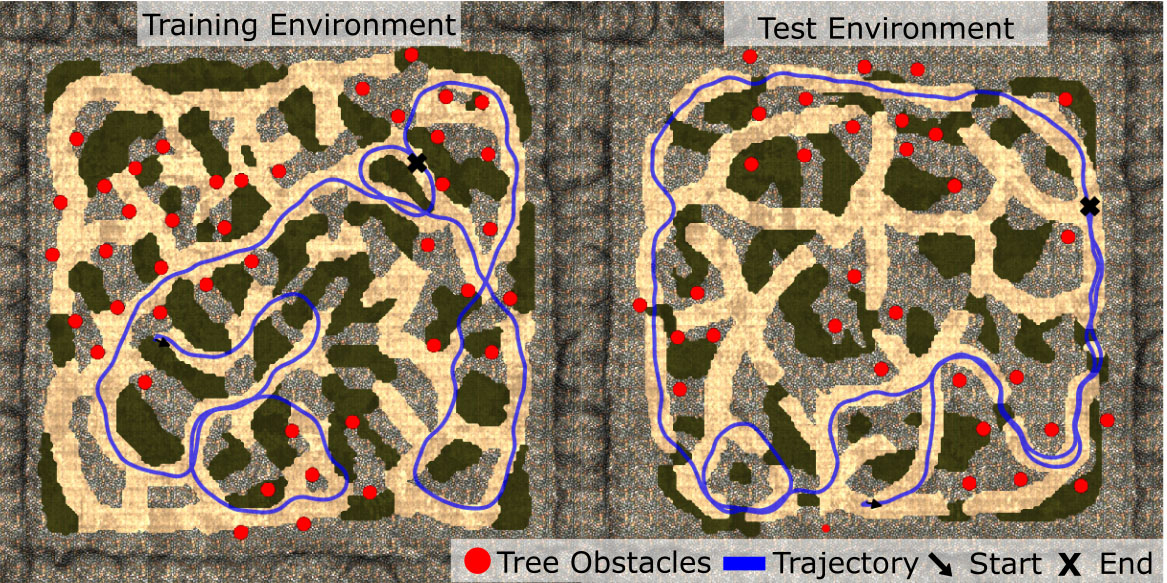}
  \caption{Map of simulation environments with sample on-policy evaluation trajectories.}
\label{fig:sim_environment_map}
\end{figure}

\vspace*{-0.05in}
We evaluated our model in off-road environments developed using an Unreal Engine simulator. Two environments were made (shown in Fig.~\ref{fig:sim_environment_map}) for the purposes of training and testing generalization, respectively. The environments were divided into sections of tall grass and tree cover to evaluate our model's ability to jointly exploit air and ground imagery inputs depending on visual obstructions in either view. A set of four terrain classes was used: $C=\langle 0: \text{smooth, } 1:\text{medium, } 2:\text{rough, } 3:\text{obstacle} \rangle$.

A simulated off-road buggy with full drivetrain physics and suspension acted as the learner. Ground and aerial cameras were configured with 90$^{\circ}$ \ac{FOV}. The aerial camera was placed 15~m above the vehicle. A control rate of 6~Hz and a predictive horizon of $H=12$ was chosen.

\subsubsection{Data Collection}

\begin{figure}[t!]
  \centering
  \includegraphics[width=0.9\linewidth]{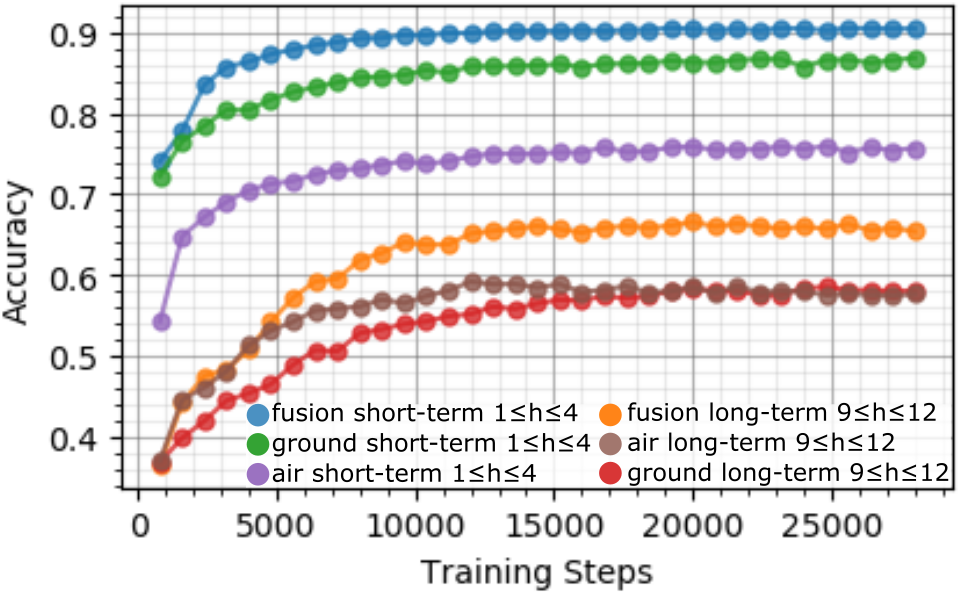}
  \caption{Simulation validation set accuracy for different models over short-term and long-term planning horizons.}
\label{fig:sim_accuracy_comparison}
\end{figure}

\begin{figure}[b!]
  \centering
  \includegraphics[width=1.0\linewidth]{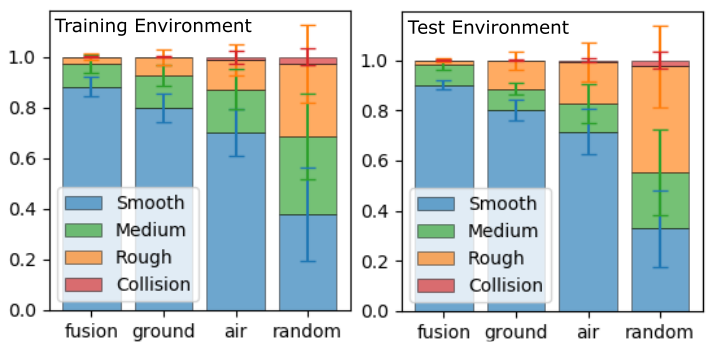}
  \caption{Percentage terrain traversed during simulation on-policy evaluation.}
\label{fig:sim_onpolicy_terrain_percentages}
\end{figure}

Data collection and training was done off-policy. Applying a random exploratory action at each timestep was found to generate trajectories with excessive stuttering and insufficient variety. Instead, a random time interval was drawn from a Gaussian distribution during which a random, uniformly sampled steering action was applied and held constant. Additional but small Gaussian noise was applied to the action at each timestep during this interval. This process was repeated iteratively.




Approximately 75,000 samples were collected, amounting to $\approx$ 3.5~hours of driving. The ground-truth terrain and collision labels were queried directly from the simulator.

\subsubsection{Off-Policy Training}
\label{sec:sim_offpolicy_training}

The network was trained on over 28,000 timesteps with a batch size of 32 and a learning rate of $\alpha=10^{-4}$. A training and validation set with a 3:1 split was used. The loss function given by Eq. \ref{eq:loss_function} was optimized with Adam ~\cite{kingma2014} and a L2 weight regularization factor of $\lambda=10^{-6}$. Training was repeated for ground-only, air-only, and fusion visual input networks. The ground-only and air-only networks are analogous to the fusion network shown in Fig.~\ref{fig:network} with the alternative camera input branches being removed.

To gain further insight into the predictive breakdown of our model, we plot the validation set accuracy divided into short-term and long-term portions of the prediction horizon $H$, as shown in Fig. \ref{fig:sim_accuracy_comparison}. 

When comparing ground versus air for short-term terrain predictions, we found that the ground view attained higher accuracy as it provided a closer, more detailed view of the immediate terrain being navigated. The short-term view of the aerial perspective was also prone to complete obstruction when tree cover was present, making it inclined to collisions. Conversely, for longer horizons, the aerial viewpoint enabled some performance gains on certain trajectories due to its increased \ac{FOV}, (apparent for turning trajectories) and its ability to see over small barriers. However, the ground view could generally see further in the forward-facing direction in the case of no obstacles or turning trajectories, making its long-term predictions preferable in other situations. Nonetheless, the fusion model outperformed both ground-only and air-only architectures, as the model learned to leverage data from both viewpoints to maximize its \ac{FOV} and counteract visual obstructions in either view. 

\subsubsection{On-Policy Evaluation}

\setlength{\belowcaptionskip}{0pt}
\setlength\tabcolsep{3pt}
\begin{table}[h]
  \centering
  \caption{Simulation average episode on-policy return ($\pm\sigma$).} \label{tab:onpolicy_traj_return}
  \vspace{0.05in}
    \begin{tabular}[t]{l|cccc}
     
     Env. & Fusion & Ground & Air & Random \\
     \hline
     Train  & 1768.4$\pm$496.1 & 1527.8$\pm$644.7 &   618.5$\pm$460.4 & 178.5$\pm$189.6 \\
     Test &   2026.3$\pm$253.2 & 1621.4$\pm$531.2   & 576.8$\pm$441.6 & 155.7$\pm$127.7 \\
     \hline
    \end{tabular}
\end{table}
\setlength{\belowcaptionskip}{-8pt}

Each trained model was evaluated on-policy for over 30 trajectories. Trajectories were terminated at the first collision or once a maximum of two minutes had elapsed. The average return (or sum of rewards) of each model is recorded in Table \ref{tab:onpolicy_traj_return}. Fig.~\ref{fig:sim_onpolicy_terrain_percentages} plots the average terrain percentage traversed by each model. Fig.~\ref{fig:sim_environment_map} shows sample trajectories generated by the fusion model. 

During on-policy evaluation, the fusion model was again found to perform best, as it learned to incorporate data from both viewpoints, thus making it robust to visual obstructions and allowing it to maximize its \ac{FOV}. The performance gains were more notable in the test environment, as the test environment had reduced tree cover (decreasing the ground model's relative performance). The fusion model attained a 90\% smooth terrain traversal rate in the test environment and a rough terrain traversal rate of only 1.83\% compared to 11.22\% of the ground-only model (6.1 times reduction).

%
%

%

\vspace*{-0.08in}
\subsection{Real-world Field Trials}
\begin{figure}[t]
\vspace{0.07in}
  \centering
  \includegraphics[width=0.95\linewidth]{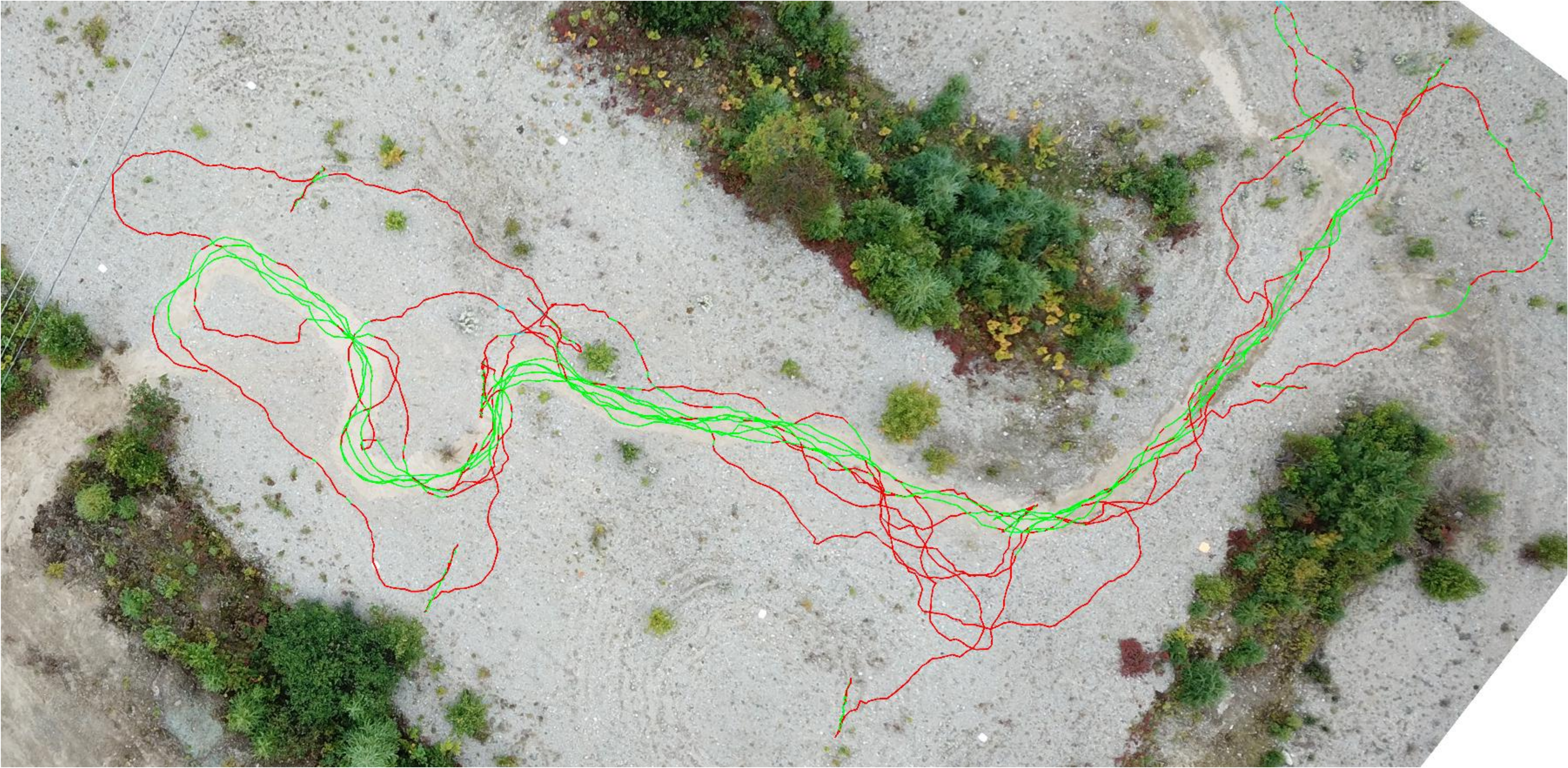}
  \caption{ An example path executed during training overlayed on the aerial image. Detected smooth terrain is shown in green, while rough terrain is shown in red.}
\label{fig:training-runs}
\end{figure}

\begin{figure}[b]
  \centering
  \includegraphics[width=1.0\linewidth]{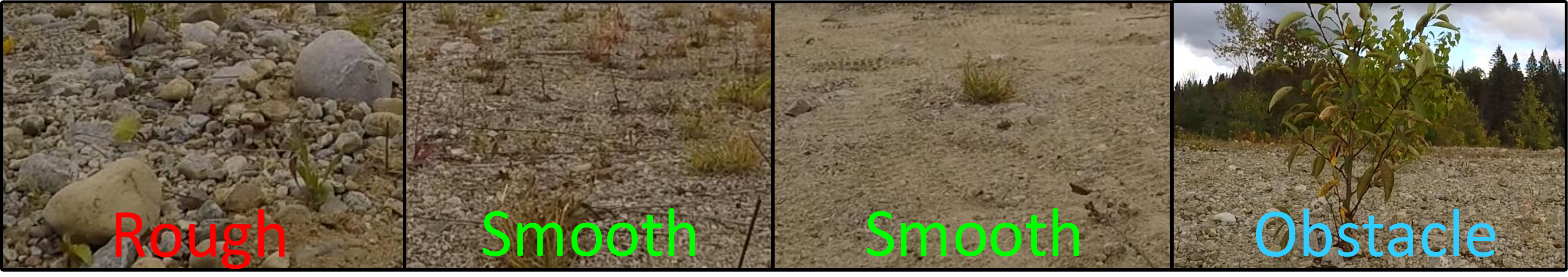}
  \caption{Example of terrain that was covered during our real-world validation experiments.}
\label{fig:three-terrains}
\end{figure}

We validated our model in a real-world, rugged, outdoor environment spanning roughly 0.25~$km^2$ using the robot system described in Sec.~\ref{sec_system_overview}. Fig. \ref{fig:training-runs} shows an example path that was executed while training our system. Three terrain types were used: $C=\{ 0: \text{smooth, } 1:\text{rough, } 2:\text{obstacle} \}$. Fig.~\ref{fig:three-terrains} shows terrain samples that were measured both qualitatively and quantitatively  (visible small and large bushes and trees were also treated as obstacles).

\subsubsection{Data Collection}
We collected joint ground and air imagery in this environment using the left camera of the ground vehicle and aerial view from a DJI Mavic Pro drone. The \ac{MAV} was used to construct a mosaic image of the experimental region, taken at an altitude of 80~m. We aligned this aerial image using four landmarks measured with RTK GPS. Training trajectories were generated consisting of tuples made from the vehicle's left camera image, corresponding aerial image patch, labeled terrain class, and steering action. Data was sampled every 0.35~m travelled instead of at fixed time interval. This sampling approach was used in an attempt to mitigate unintentional speed changes due to factors such as draining battery voltage. A planning horizon of $H=16$ was selected. 

Terrain labeling followed a purely self-supervised approach. We measured the \ac{RMS} linear acceleration values in the direction normal to the ground plane reported by the robot's on-board \ac{IMU} over a short window of 20 60~Hz samples. A frequency representation of this short window signal was obtained by taking the Fourier transform and then the magnitude component was put into a 15 bin histogram divided by frequency. The histogram was used to make a feature vector processed by K-Means clustering to generate labels. The on-board Lidar was used to label obstacles and pre-emptively stop the vehicle prior to physical collision.

\begin{figure}[t!]
\vspace{0.07in}
  \centering
  \includegraphics[width=0.9\linewidth]{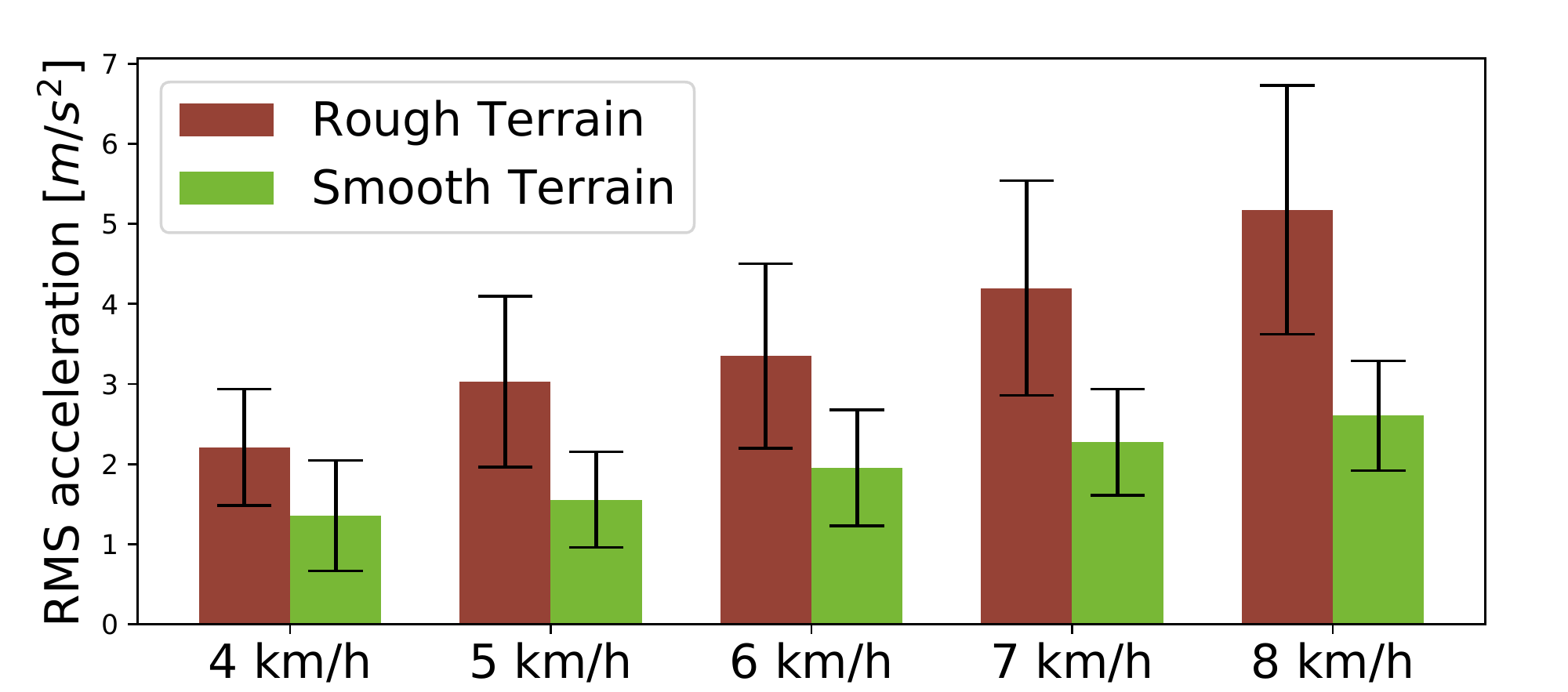}
  \caption{RMS linear acceleration versus speed and terrain type for rough and smooth classes.}
\label{fig:rms_vs_speed}
\end{figure}

\begin{figure}[b]
  \centering
  \includegraphics[width=1\linewidth]{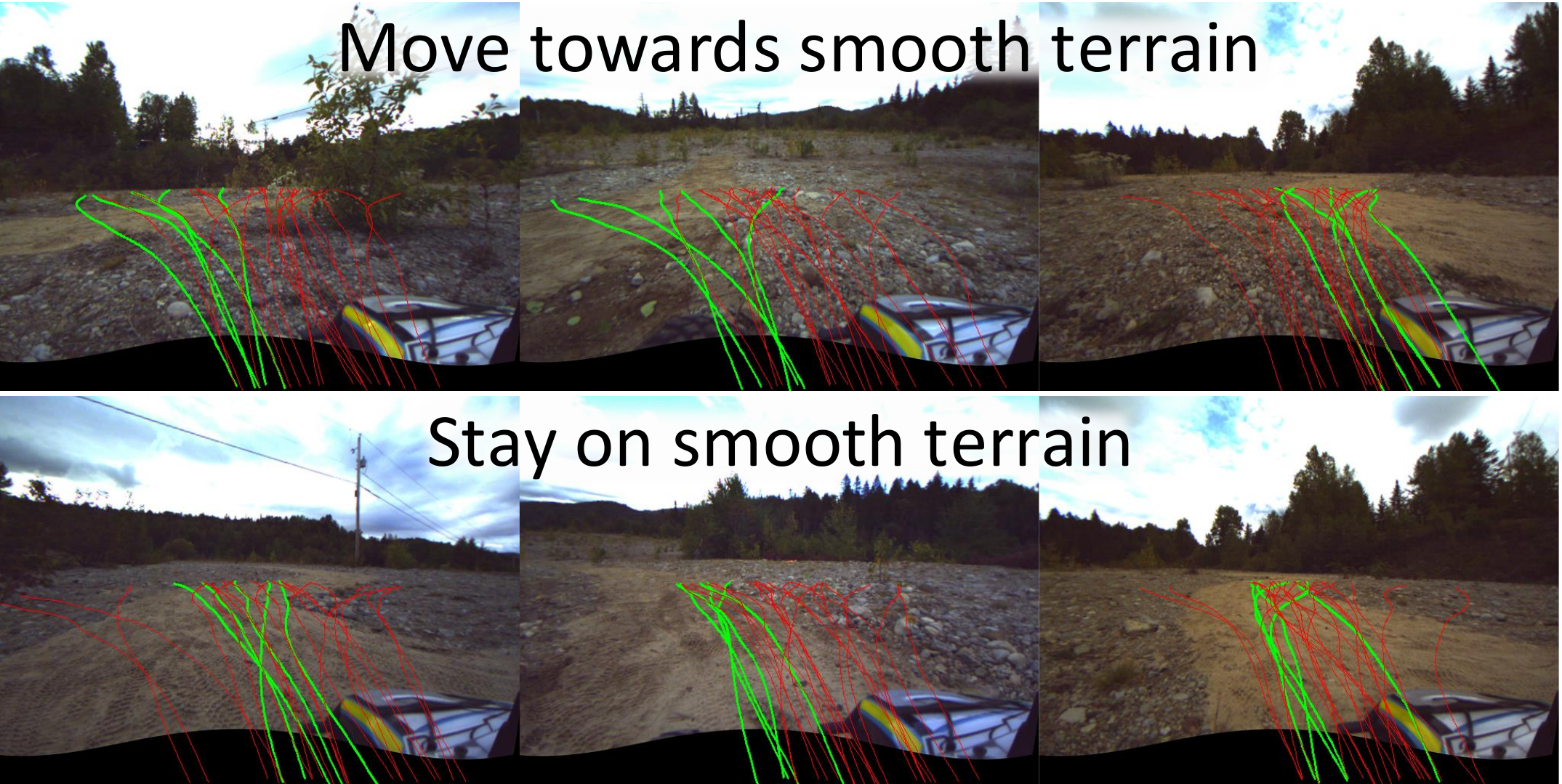}
  \caption{Sample action predictions for moving towards and staying on smooth terrain. High reward predictions are shown in green while low reward predictions are shown in red.}
\label{fig:action_predictions}
\end{figure}


We also experimented with how our assigned roughness scores varied with speed. Fig. \ref{fig:rms_vs_speed} shows that the \ac{RMS} metric became more noticeable at higher speeds, but actions were more uncertain due to violent vibrations and impacts. In our experiments, we selected 6 km/h speed, which still provided a good distinction between terrain classes.


The collected trajectories amounted to approximately 5.25~km or $\approx~15,000$ data samples.

\subsubsection{Training and Evaluation}

Despite a relatively small dataset, we obtained reasonable performance after $\approx$ 8,000 training steps. The prediction accuracy using the forward ground camera alone ranged from 78\% to 60\% over horizon intervals of 1 to 16, respectively. For actions that were generally forward in direction, there was minimal improvement using the aerial image. However, we found a consistent  improvement in accuracy of $\approx$10\% when incorporating aerial imagery for trajectories possessing a change in angle of 45$^{\circ}$ or more. 
Fig. \ref{fig:action_predictions} shows sample path predictions of the trained model navigating towards smooth terrain.

\section{CONCLUSIONS}
\label{sec_conclusion}

This paper has described the first application of a deep terrain predictive model using ground-air image fusion and self-supervised training to learn a smooth, collision-free navigation policy in rugged outdoor environments. Through field deployments and extensive simulations, we have shown that our model plans collision-free trajectories that maximize smooth terrain with limited training data requirements. Our key contributions include the use of IMU self-supervision as a proxy for driveability and the combined use of aligned aerial images and first-person on-board images. This multi-modal input strengthens the predictive capabilities of the model by increasing the available field of view and making it robust to visual obstructions.




\begin{acronym}
\acro{CNN}{Convolutional Neural Network}
\acro{RNN}{Recurrent Neural Network}
\acro{VPN}{Value Prediction Network}
\acro{GCG}{General Computation Graph}
\acro{DNN}{Deep Neural Network}
\acro{FOV}{Field of View}
\acro{ORB}{Oriented FAST and Rotated BRIEF}
\acro{DSO}{Direct Sparse Odometry}
\acro{MAV}{Micro Air Vehicle}
\acro{UAV}{Unmanned Aerial Vehicle}
\acro{IMU}{Inertial Measurement Unit}
\acro{ROS}{Robot Operating System}
\acro{GPS}{Global Positioning System}
\acro{RTK}{Real-Time Kinematic}
\acro{RMS}{Root Mean Square}
\acro{SVM}{Support Vector Machine}
\acro{LSTM}{Long Short-Term Memory}
\end{acronym}

\maxpage[{\color{red}\Large Main text is too long.}]{6}

\pagebreak

\bibliographystyle{IEEEtran}
\bibliography{MyBSTcontrol, icra2020_offroad_driving, library}

\end{document}